\documentclass[11pt,a4paper]{article}
\usepackage[hyperref]{emnlp2020}
\usepackage{times}
\usepackage{latexsym}

\usepackage{microtype}

\usepackage{amsfonts}
\usepackage{amsmath}
\usepackage{booktabs}
\usepackage{multirow}
\usepackage{graphicx}
\usepackage{enumitem}
\usepackage{textcomp}
\usepackage{booktabs}
\usepackage{subfigure}
\aclfinalcopy % Uncomment this line for the final submission

\newcommand*{\email}[1]{#1}

\title{``What Do You Mean by That?''\\ A Parser-Independent Interactive Approach for Enhancing Text-to-SQL}

\author{Yuntao Li\textsuperscript{1}\thanks{\ \ Work done during an internship at Microsoft Research.}~,
Bei Chen\textsuperscript{2},
Qian Liu\textsuperscript{3}$^*$,
Yan Gao\textsuperscript{2},
Jian-Guang Lou\textsuperscript{2},
Yan Zhang\textsuperscript{1},
Dongmei Zhang\textsuperscript{2}\\
\textsuperscript{1}{Department of Machine Intelligence, Peking University, Beijing, China}\\
\textsuperscript{2}{Microsoft Research, Beijing, China}; 
\textsuperscript{3}{Beihang University, Beijing, China}\\
\textsuperscript{1}\email{\{li.yt, zhyzhy001\}@pku.edu.cn}; 
\textsuperscript{3}\email{qian.liu@buaa.edu.cn}\\
\textsuperscript{2}\email{\{beichen, yan.gao, jlou, dongmeiz\}@microsoft.com}
}

\date{}

\begin{document}
\maketitle
\begin{abstract}
In Natural Language Interfaces to Databases systems, the text-to-SQL technique allows users to query databases by using natural language questions. Though significant progress in this area has been made recently, most parsers may fall short when they are deployed in real systems. One main reason stems from the difficulty of fully understanding the users' natural language questions. In this paper, we include human in the loop and present a novel parser-independent interactive approach (PIIA) that interacts with users using multi-choice questions and can easily work with arbitrary parsers. Experiments were conducted on two cross-domain datasets, the WikiSQL and the more complex Spider, with five state-of-the-art parsers. These demonstrated that PIIA is capable of enhancing the text-to-SQL performance with limited interaction turns by using both simulation and human evaluation. 
\end{abstract}

\section{Introduction}

The past few years have witnessed a burgeoning interest in the study of text-to-SQL, the essential technique for Natural Language Interfaces to Databases (NLIDB) systems \cite{guo2019towards,hwang2019comprehensive,he2019x,bogin2019global,bogin2019representing}. 
By converting natural language (NL) questions into executable forms (i.e., Structured Query Language or SQL), text-to-SQL parsers relieve users from the burden of learning about techniques behind the queries.
Though significant progress has been made in this field, most parsers are still less than desirable when deployed in real NLIDB systems. As users are not experts in database querying, a central challenge for the parsers is to fully understand the users' NL questions.

Since users are who know the questions best, interacting with them has been seen as a promising way to tackle the above challenge in real NLIDB systems.
Early works tried to get users involved in checking SQL queries \cite{li2014constructing,iyer2017learning,yaghmazadeh2017sqlizer}, which are impracticable in real systems, as they can only succeed if users have a very good knowledge of SQL. In another attempt to involves users,  \citet{gur2018dialsql} proposed to interact with non-expert users by multi-choice questions. However, this approach is designed for relatively simple scenarios and cannot be easily applied to more complex ones. More recently, \citet{yao2019model} took an important step forward by measuring uncertainty of neural-based parsers and altering the behavior of them.
However, as far as we know, most parsers in real systems are equipped with elaborate rules instead of using fully neural methods \cite{dhamdhere2017analyza,gliozzo2013semantic,lai2014systems}. Moreover, in some situations, parsers are supplied by third parties, making it impossible to alter them. Therefore, assuming parsers are a black box, it is indispensable to conduct research on an interactive approach for enhancing the text-to-SQL technique in complex scenarios.

In this paper, we propose a \textbf{P}arser-\textbf{I}ndependent \textbf{I}nteractive \textbf{A}pproach (PIIA) to interact with human users and help parsers better understand NL questions.
To achieve this goal, we devised three modules: 
(1) \emph{Error Locator} employs an alignment method to help parsers locate uncertain tokens in the NL questions.
(2) \emph{Question Generator} designs multi-choice questions in natural language for users, which offers a pleasant interactive experience.
(3) \emph{NL Modifier} rewrites the NL questions according to the users' feedback and produces more legible questions to facilitate downstream parsing. Our major contributions are:
\begin{itemize}[leftmargin=*]\setlength\itemsep{-0.3em}
    \item We propose a novel interactive approach, named PIIA, to enhance the text-to-SQL for complex SQL queries in a cross-domain scenario.
    \item The interaction process in PIIA is user-friendly that asks multi-choice questions and reduces the number of questions as much as possible. 
    \item PIIA is designed as a parser-independent approach that can easily collaborate with arbitrary base parsers and be deployed in real systems.
    \item We conduct a series of experiments with five base parsers on two large cross-domain datasets that demonstrate the effectiveness of PIIA by using both simulation and human evaluation.
\end{itemize}

\begin{figure}
    \centering
    \includegraphics[width=1\linewidth]{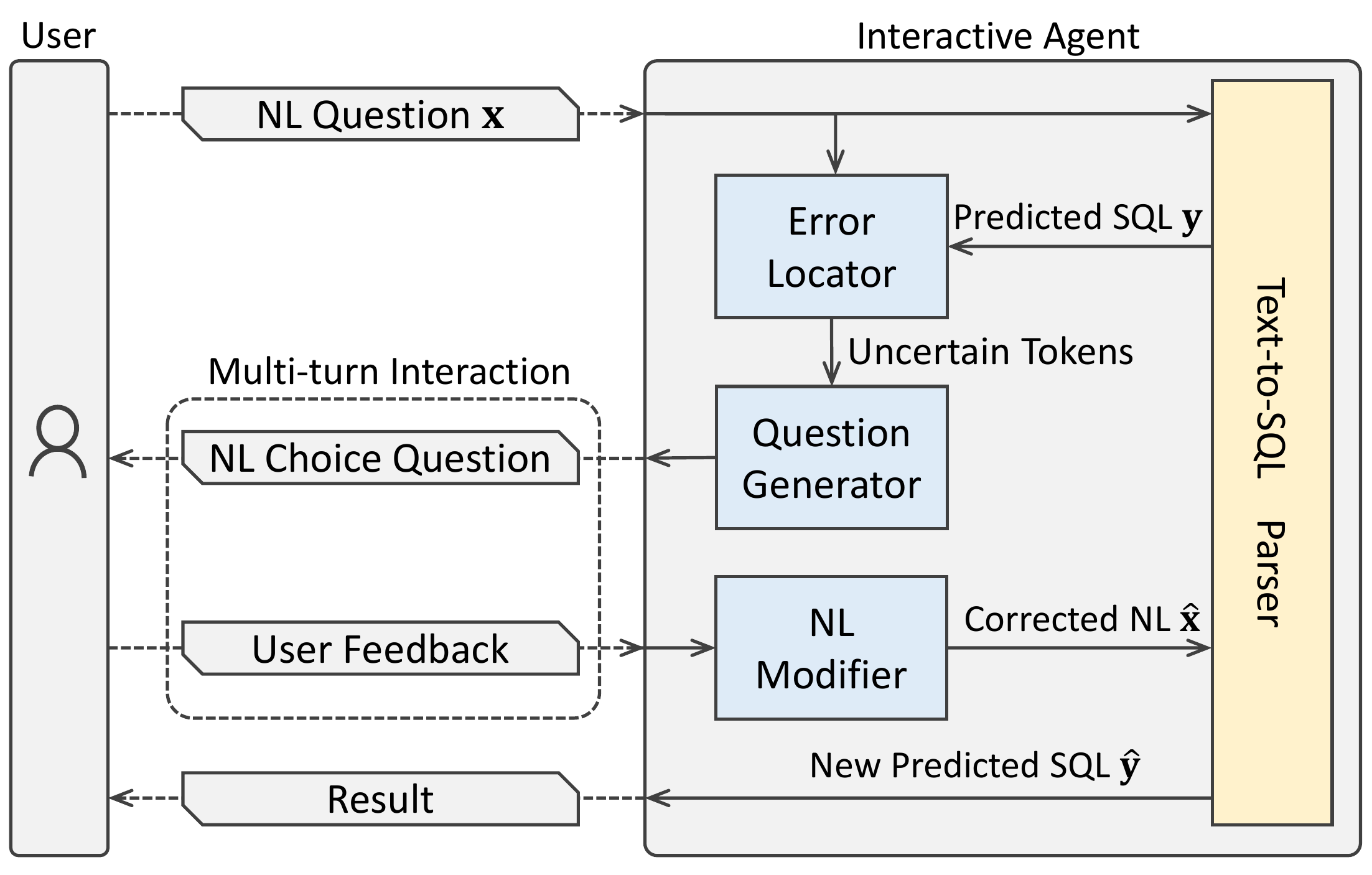}
    \caption{The schema of PIIA, consisting of Error Locator, Question Generator and NL Modifier.}
    \label{fig:overview}
\end{figure}

\section{Methodology Overview}
\label{sec:overview}

While querying databases in an NLIDB system, users pose a natural language question that is denoted as $\mathbf{x}$. The text-to-SQL parser takes $\mathbf{x}$ as input and predicts a SQL query, which is denoted as $\mathbf{y}$. The system then executes the predicted SQL and returns the result. 
As mentioned, users are not experts in database querying, so they may pose natural language questions with inexplicit expressions.
To better understand the difficulties caused by inexplicit expressions, we carefully analyzed 300 mistakes made by IRNet \cite{guo2019towards}, one of the state-of-the-art parsers on the Spider dataset \cite{yu2018spider}. We found that in 47.3\% of the cases the parser couldn't understand database-related information, such as table and column names as well as values. 
Thus, we build PIIA upon parsers that can interactively revise inexplicit expressions in $\mathbf{x}$ with the help of users' feedback, thus enhancing the performance of text-to-SQL.

Our proposed PIIA, shown schematically in Figure \ref{fig:overview}, works between the user and the text-to-SQL parser and it consists of three modules, Error Locator, Question Generator and NL Modifier. 
After receiving $\mathbf{x}$ and $\mathbf{y}$, the Error Locator helps the parser find a set of uncertain tokens in $\mathbf{x}$.
For each uncertain token, the Question Generator creates a natural language multi-choice question. After interactively asking the user all the multi-choice questions, the PIIA agent collects all the answers (i.e., the user's selections). Then, the NL Modifier corrects $\mathbf{x}$ based on the answers and obtains a more legible question $\mathbf{\hat{x}}$. Finally, by feeding modified question $\mathbf{\hat{x}}$ into the text-to-SQL parser, we can get a new predicted SQL query $\mathbf{\hat{y}}$. Compared to $\mathbf{x}$, $\mathbf{\hat{x}}$ combines user's feedback and contains clearer semantics. Hence $\mathbf{\hat{y}}$ is likely more accurate than $\mathbf{y}$. Note that, our PIIA agent will not read the contents of the databases (i.e., values) due to privacy concerns. Details of the three modules are presented in Sections \ref{sec:errorlocation}, \ref{sec:questiongeneration} and \ref{sec:nlmodifier}.

\begin{figure}
    \centering
    \includegraphics[width=0.48\textwidth]{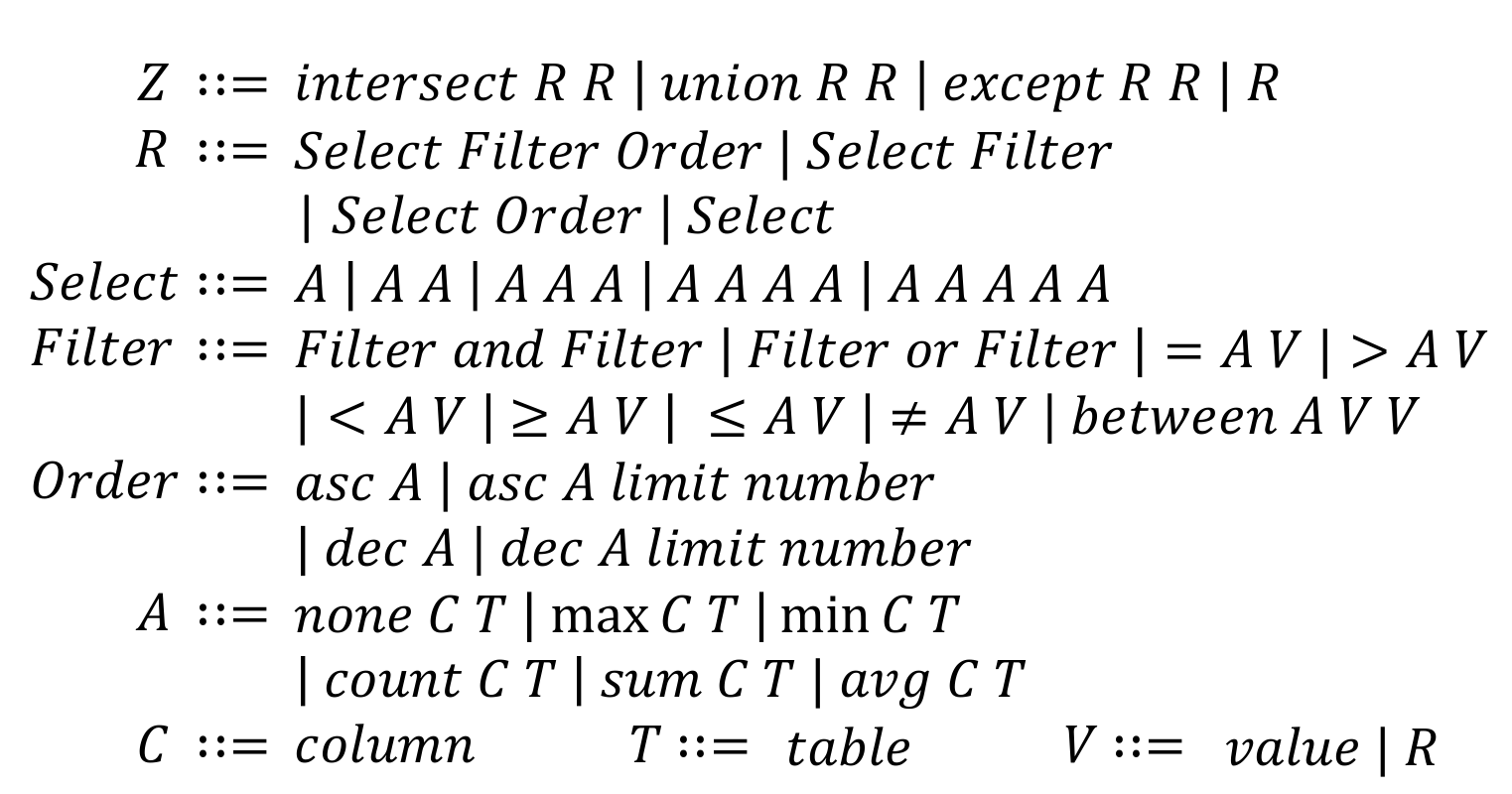}
    \caption{The grammar of the intermediate language. In a specific database, $column$ refers to distinct column names while $table$ comprises several table names and $value$ indicates the value tokens expressed by the user.}
    \label{fig:grammar}
\end{figure}

\begin{figure*}[ht]
    \centering
    \includegraphics[width=0.95\textwidth]{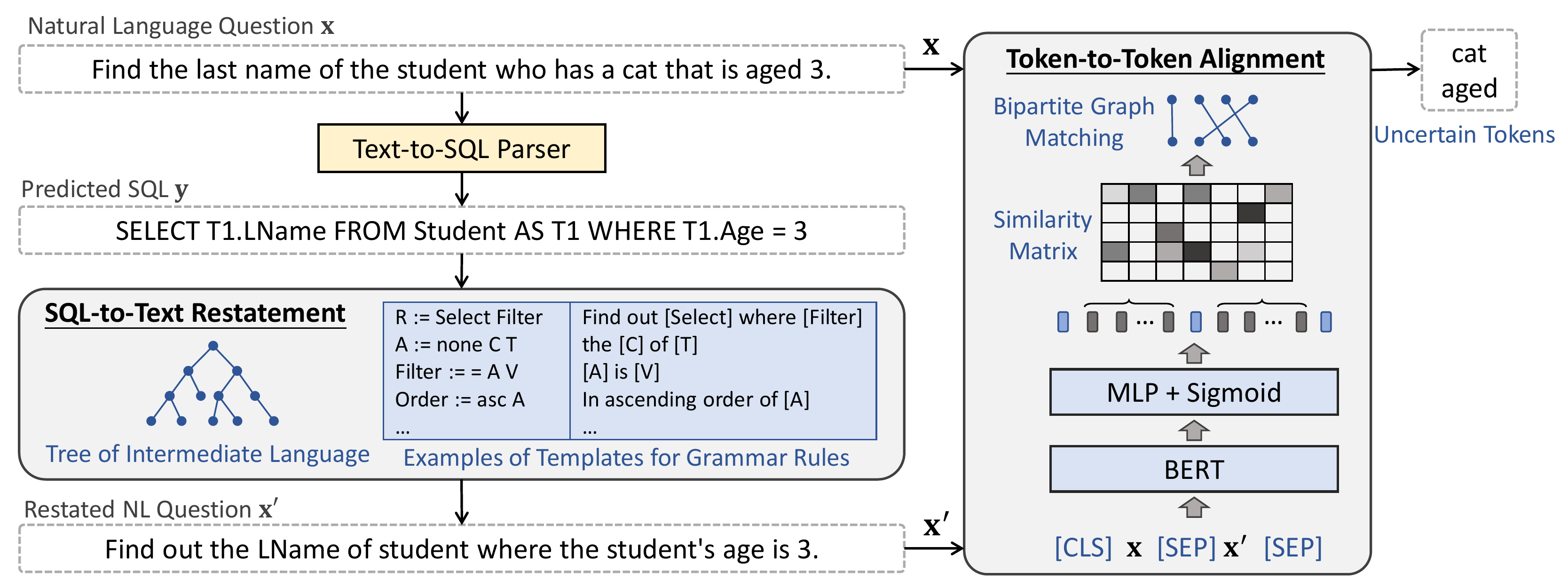}
    \caption{Illustration of Error Locator with a real case from IRNet on the Spider dataset. The NL question $\mathbf{x}$ is parsed to SQL $\mathbf{y}$, and then converted to restated NL question $\mathbf{x}'$ via SQL-to-text restatement. Then, a token-to-token alignment similarity matrix between $\mathbf{x}$ and $\mathbf{x}'$ is computed to detect uncertain tokens (i.e., cat and aged).}
    \label{fig:errorlocator}
\end{figure*}

\section{Error Locator}
\label{sec:errorlocation}

The goal of Error Locator is to detect the tokens in $\mathbf{x}$ that are hardly understood by the parser. These are called \emph{Uncertain Tokens}, and in most cases are related to database information, such as table and column names as well as values. 
When the parser fails to understand them, the uncertain tokens will be mistranslated or ignored.
Since the parser is a black box, the only information we can get from it is the predicted SQL. Hence, we devise a method to compare the NL question $\mathbf{x}$ with the predicted SQL $\mathbf{y}$. All the informative tokens in $\mathbf{x}$ may align with the corresponding tokens in $\mathbf{y}$, while the unaligned tokens in $\mathbf{x}$ are extracted as uncertain tokens. To this end, we firstly restate the predicted SQL $\mathbf{y}$ to an NL question $\mathbf{x}'$ through \emph{SQL-to-Text Restatement}, and then perform \emph{Token-to-Token Alignment} between $\mathbf{x}$ and $\mathbf{x}'$.

\subsection{SQL-to-Text Restatement}

Compared to aligning an NL question with a SQL query, the alignment of two NL questions is more reasonable because it utilizes a similar linguistic structure and can make better use of pre-trained models (e.g., BERT). Thus, before the alignment, we restate the predicted SQL $\mathbf{y}$ into a natural language question $\mathbf{x}'$. Previous work has proposed sequence-to-sequence based methods \cite{guo2018question} that cannot ensure the restatement correctness. In contrast, we carefully design a template-based SQL-to-text method that solves this problem because the restatement correctness and integrality are critical for the alignment process.

Since SQL is execution-oriented, its clauses are about database operations. Some of the clauses may not be expressed in the users' NL questions, such as \texttt{GROUPBY} and \texttt{JOIN}. To bridge the gap between natural language and SQL, we design an intermediate language that is inspired by \citet{guo2019towards} and whose grammar is shown in Figure \ref{fig:grammar}. The predicted SQL can be easily converted into the intermediate language, which can be naturally represented by a hierarchical tree structure. Each tree node corresponds to a grammar rule. For each kind of grammar rule, we design a few natural language templates to describe it. Thus, we can recursively convert the tree into a natural language question. As for nested SQL queries, we utilize subordinate clauses with ``that/which'' to handle the subqueries. The SQL-to-text restatement is depicted on the left of Figure \ref{fig:errorlocator}, along with a concrete example. We also give some examples of templates in the figure. Finally, the restated $\mathbf{x}'$ has the same semantics of SQL $\mathbf{y}$ and is independent of database internal operations.

\subsection{Token-to-Token Alignment}

Given the user NL question $\mathbf{x}$ and the restated NL question $\mathbf{x}'$, we perform the token-to-token alignment between them and find out the uncertain tokens in $\mathbf{x}$. As shown on the right of Figure \ref{fig:errorlocator}, we adopt BERT \cite{devlin2018bert} as the encoder. BERT is pre-trained on a large corpus and equipped with the ability to encode sentences on the basis of contextual information. 
The input of BERT is the concatenation of the two questions with ``[CLS]" and ``[SEP]". 
Each token obtains an output vector from BERT, which is fed into a trainable Multi-Layer Perceptron (MLP) layer to further distill useful information.
Assuming that $\mathbf{x}$ has $N$ tokens and $\mathbf{x}'$ has $M$ tokens, we can denote that $\mathbf{x}=(x_1,x_2,\dots,x_N)$ and $\mathbf{x}'=({x}_1',{x}_2',\dots,{x}_M')$. The output embeddings of tokens in $\mathbf{x}$ and $\mathbf{x}'$ can respectively be denoted by
\begin{equation}
\begin{aligned}
H&=(\mathbf{h}_1,\mathbf{h}_2,\dots,\mathbf{h}_N)&&\in \mathbb{R}^{d\times N},\\
U&=(\mathbf{u}_1,\mathbf{u}_2,\dots,\mathbf{u}_M)&&\in \mathbb{R}^{d\times M},
\end{aligned}
\end{equation}
where $d$ is the output embedding size. 

Based on $H$ and $U$, we employ the cosine similarity to derive a token-level similarity matrix $A\in \mathbb{R}^{N \times M}$, where each entry $A_{nm}$ indicates the similarity between $x_n$ and ${x}_m'$:
\begin{equation}
A_{nm} = \frac{\mathbf{h}_n^{\top} \cdot \mathbf{u}_m}{||\mathbf{h}_n|| \cdot ||\mathbf{u}_m||}.
\end{equation}
The alignment between $\mathbf{x}$ and $\mathbf{x}'$ can be obtained using the similarity scores in $A$. After removing stop words and words from SQL-to-text restatement templates, we regard the similarity scores as the weights of a bipartite graph and apply the Hungarian maximum matching algorithm to find an optimized token-level one-to-one alignment. Finally, the tokens in $\mathbf{x}$ with alignment scores less than the threshold $p$ are extracted as uncertain tokens.
It is also important to note that a schema-aware post-processing is operated on these scores. Since the tokens that appear more than once in the database schema may confuse the alignment process, the post-processing aims to give addition bias and help Error Locator detect potential uncertain tokens\footnote{We lower the alignment score of a token (denoted as $S$) by the number of its occurrences (denoted as $C$) in the database schema (the score is lowered to $S/C$).}.

\subsection{Training Process}
Since the annotations of token-to-token alignment are not available, fully supervised learning is infeasible. Inspired by the work of \citet{legrand2016neural}, we solve this problem by leveraging negative sampling to generate training data and adopt a weakly supervised training strategy. 
Concretely, we collect several pairs $(\mathbf{x},\mathbf{x}'_{pos})$. $\mathbf{x}$ is the user NL question and $\mathbf{x}'_{pos}$ is the corresponding positive restated NL question restated from the ground truth SQL of $\mathbf{x}$. For each $(\mathbf{x},\mathbf{x}'_{pos})$ pair, we generate negative restated NL questions $\mathbf{x}'_{neg}$ in two ways: 
random sampling, where we randomly pick an $\mathbf{x}'$ from other pairs as $\mathbf{x}'_{neg}$;  
and perturbed sampling, where we generate an $\mathbf{x}'_{neg}$ by replacing column names or/and value tokens in $\mathbf{x}'_{pos}$. 
The random samples are vastly different from $\mathbf{x}'_{pos}$, and the common tokens in positive and negative samples are uninformative (e.g., stop words). This kind of $\mathbf{x}'_{neg}$ helps distinguish the informative and uninformative tokens. The perturbed samples have the same uninformative tokens with $\mathbf{x}'_{pos}$, and help model to focus on the alignment of informative tokens. We generate $50$ random samples and $50$ perturbed samples for each $(\mathbf{x},\mathbf{x}'_{pos})$ pair.

By generating negative samples, we obtain the training data composed of triples $(\mathbf{x},\mathbf{x}'_{pos},\mathbf{x}'_{neg})$. The intuition behind the weakly supervised training is that $\mathbf{x}$ is more similar to $\mathbf{x}'_{pos}$ than $\mathbf{x}'_{neg}$. We measure the sentence-level similarity by averaging the token-level similarities:
\begin{equation}
s(\mathbf{x},\mathbf{x}')=\frac{1}{N}\sum_{n=1}^N \max_{m=1}^M A_{nm}.
\end{equation}
Then, our goal is to increase $s(\mathbf{x},\mathbf{x}'_{pos})$ and decrease $s(\mathbf{x},\mathbf{x}'_{neg})$. We employ hinge loss to maximize the margin of the two scores, which is also accompanied by $L^1$-norm on two corresponding similarity matrices to make them sparse. The loss function is:
\begin{equation}
\begin{aligned}
\!\!\!\!L =& \max\!\left( 0,m-(s(\mathbf{x},\mathbf{x}'_{pos})-s(\mathbf{x},\mathbf{x}'_{neg}))\right) \\
\!\!\!\!&+ \lambda \left( \vert A_{pos}\vert_1 + \vert A_{neg}\vert_1\right),
\end{aligned}
\end{equation}
Where $m$ is the margin and $\lambda$ balances the hinge loss and the $L^1$-norm.

\begin{figure}
    \centering
    \includegraphics[width=0.48\textwidth]{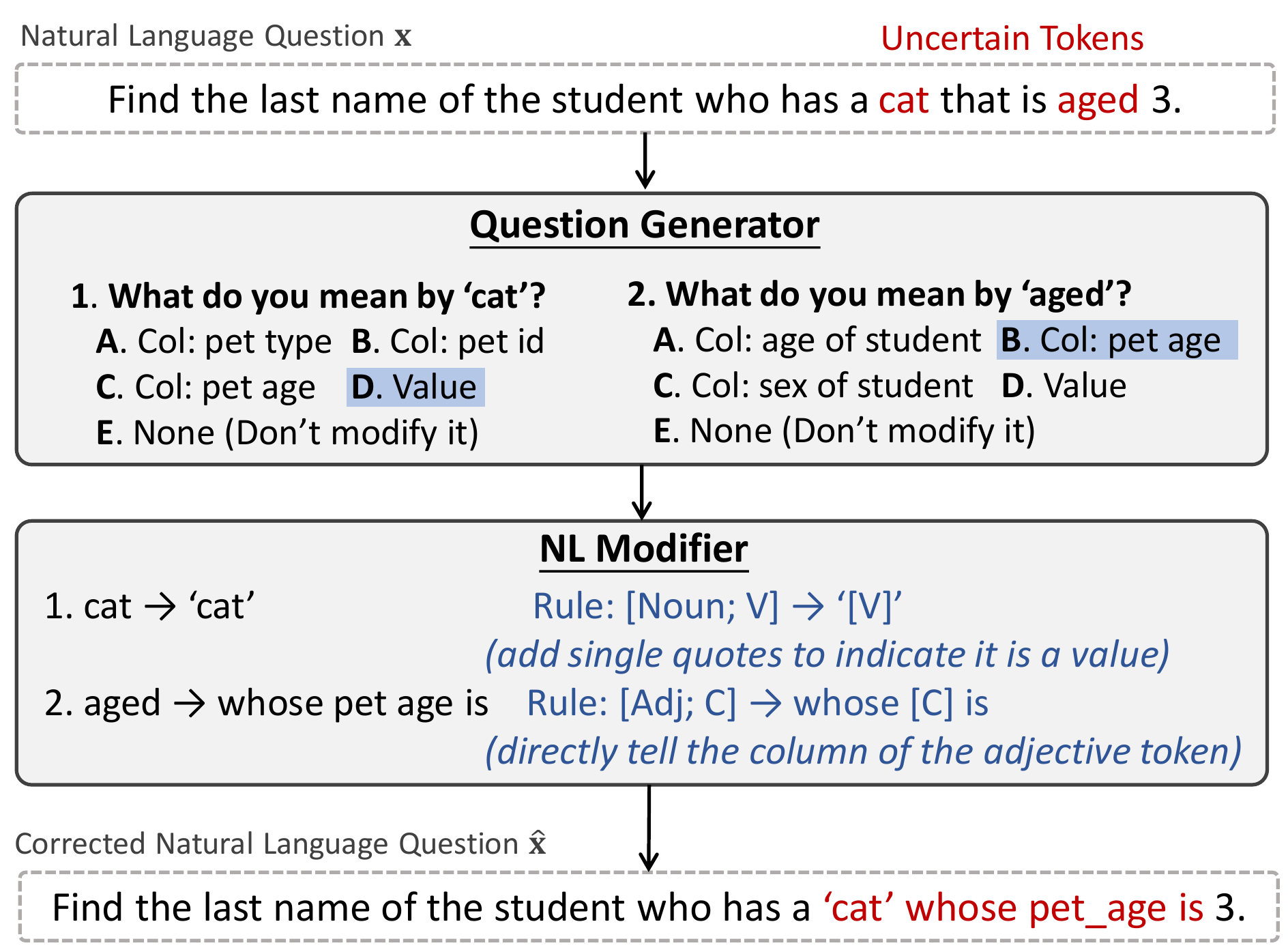}
    \caption{Question Generator and NL Modifier: an example. Shaded options in the multi-choice questions are selected by the user.}
    \label{fig:question}
\end{figure}

\section{Question Generator}
\label{sec:questiongeneration}

For each uncertain token detected by the Error Locator, the PIIA agent interacts with the user to get a more explicit explanation. Instead of asking the user to explain the uncertain token directly, we provide an NL multi-choice question. As users are non-expert and unfamiliar with database operations, simply picking an option is more natural and friendly. Thus, the Question Generator is designed to generate a multi-choice question for each uncertain token\footnote{During interaction, after getting an answer to a multi-choice question, we check the remaining uncertain tokens. If an uncertain token exists in the answer, we delete the corresponding multi-choice question to avoid repeating.}, i.e., ``What do you mean by that?'' as shown in Figure \ref{fig:question}.

To make a multi-choice question, a set of candidate options are generated. As analyzed in Section \ref{sec:overview}, most of the uncertain tokens are related to database information. Thus, for each uncertain token in an NL question, we find out the corresponding database and add all the column and table names into the candidate set. Additionally, we observe that some uncertain tokens are about aggregation operations, so we also add the aggregation operations into the candidate set, such as \emph{min}, \emph{max} and \emph{sum}. As a result, the set is quite large, especially for complex databases. For example, the set size can be larger than 40 in the Spider dataset. 

Hence, we devise a ranking method to find out the options with the highest correlations to the uncertain token. Concretely, for each candidate option (denoted by $\mathbf{w}$), we calculate its similarity score with the uncertain token (denoted by $z$). As each candidate option is a span with one or more tokens, we adopt both lexical and semantic similarities. $\mathbf{w}$ and $z$ are pre-processed with lemmatization. Then the Jaccard distance between them is computed for lexical similarity, which is the number of common tokens divided by the number of unique tokens in them. For semantic similarity, we present each token as an embedding vector (i.e., GloVe \cite{pennington2014glove}) and employ the Euclidean distance between $\mathbf{w}$ and $z$. The embedding vector of a span is the average embedding over tokens in the span. All the candidate options are ranked by the summation of these two similarity scores, and three of them are picked as options. Additionally, we add two more options, \emph{Value} and \emph{None}, to each multi-choice question. \emph{Value} indicates that $z$ is related to a value in the database. \emph{None} means that either the token does not need modification or all the other options are not related. The \emph{None} option is essential as it prevents uncertain tokens from being modified unexpectedly. This alleviates the need for Error Locator to make exact error detection and ensures a higher recall rate.
Following the example in Figure \ref{fig:errorlocator}, we show the question generation process in Figure \ref{fig:question}. As we can observe, the multi-choice questions are easy to be understood by non-expert users, and the provided options are reasonable. After interacting with users, the PIIA agent gets the information that ``cat" is a value, and ``aged" indicates the column ``pet\_age".

\section{NL Modifier}
\label{sec:nlmodifier}

The last module of PIIA is the NL Modifier, which corrects NL questions with the users' feedback, i.e., how users answer the multi-choice questions. The most straightforward way is to directly replace the uncertain tokens with the selected options. However, it is not always reasonable. Since the uncertain tokens can be not only nouns but also verbs or adjectives, directly replacing verbs or adjectives may cause incoherence. To avoid this problem, we carefully design several modifier rules according to different POS tags, option types, and user NL question contexts. As mentioned in Section \ref{sec:questiongeneration}, there are four option types: column name, table name, aggregation, and value. A concrete example is shown in Figure \ref{sec:questiongeneration}, where we also list the modifier rules that are applied. Since the noun ``cat" is selected as a value, the single quotes are added. It is easier for the parser to recognize it as a value because values are always equipped with single quotes in the dataset. Moreover, as there are multiple columns about age in the database, the adjective ``aged" is modified to ``whose pet\_age is" to make the column name easier to identify. Finally, the corrected NL question $\mathbf{\hat{x}}$ is fed into the text-to-SQL parser. More modifier rules and examples are shown in Table \ref{fig:more_case}.

PIIA is designed to modify the user NL questions instead of the predicted SQLs. Modifying the predicted SQLs is more straightforward but impracticable. We conducted several surveys and found non-expert users had difficulty giving high-quality responses to modify SQLs directly, even for simple SQLs. Note that, the uncertain tokens found by PIIA are the unaligned tokens in NL questions, so the users' feedback to uncertain tokens cannot be used to modify the corresponding SQLs.

\section{Experiments}

In this section, we firstly introduce the experimental setup. Then we assess PIIA by using both simulation and human evaluation, and finally we perform a closer analysis of PIIA.

\subsection{Experimental Setup}

We conduct experiments on two cross-domain text-to-SQL datasets with five base parsers\footnote{Code available at \url{https://github.com/microsoft/ContextualSP}.}.

The \textbf{WikiSQL} dataset \cite{zhongSeq2SQL2017} collects 24,241 cross-domain single-table databases from Wikipedia and contains 80,654 hand-annotated pairs of NL questions and SQL queries. The SQL queries are relatively simple with only \texttt{SELECT} and \texttt{WHERE} clauses. Two parsers are selected: (1) \textbf{SQLova} \cite{hwang2019comprehensive}, currently the best open-sourced parser on WikiSQL, uses table-aware and context-aware representations of questions to generate SQL queries. (2) \textbf{SQLNet} \cite{xu2017sqlnet} applies sequence-to-set prediction and employs a sketch-based approach to predict SQL queries. 
We report our PIIA results on the test set, which contains 15,878 samples.

The \textbf{Spider} dataset \cite{yu2018spider} is a human-labeled text-to-SQL dataset that consists of 10,181 NL questions and 5,693 unique complex SQL queries on 200 databases with multiple tables. It covers 138 different domains and is much more complex than the WikiSQL dataset because it has a greater number of complex questions and nested SQL queries. Three parsers are selected: (1) \textbf{IRNet} \cite{guo2019towards}, currently the state-of-the-art open-sourced parser on Spider, employs the coarse-to-fine framework \cite{dong2018coarse} and designs an intermediate language. (2) \textbf{IRNet+BERT} takes BERT as NL encoder to enhance the performance of basic IRNet. (3) \textbf{SyntaxSQLNet} \cite{yu2018syntaxsqlnet} employs a SQL specific syntax tree based decoder and table-aware column attention encoders. 
The test set is not publicly available, so we evaluate PIIA on the development set, which contains 1,034 samples.

In Error Locator, the similarity threshold is set to be the average score of all the $(\mathbf{x}, \mathbf{x}')$ pairs in the training triples $\mathcal{X}=\{(\mathbf{x},\mathbf{x}'_{pos},\mathbf{x}'_{neg})\}$ as follows:
$$p=\frac{1}{2|\mathcal{X}|} \sum_{(\mathbf{x},\mathbf{x}'_{pos},\mathbf{x}'_{neg})\in\mathcal{X}} \!\!\!\left(s(\mathbf{x},\mathbf{x}'_{pos})+s(\mathbf{x},\mathbf{x}'_{neg})\right).$$ 
This threshold score is generally lower than alignment scores of certain tokens and higher than that of uncertain tokens, which can help to distinguish certain and uncertain token alignment. For hyper-parameters, we set $m=1$ and $\lambda=0.5$. As for the NL Modifier, the NLTK pos tagging model \cite{loper2002nltk} is employed for pre-processing.

\subsection{Simulation Evaluation}

We build a simulator to interact with the PIIA agent, which aims to give ideal selections for multi-choice questions. We report the results achieved by five base parsers.

\paragraph{Simulator}
The simulator chooses options on behalf of the real user.
Given an NL question with $T$ uncertain tokens, PIIA asks a multi-choice question with $K$ options for each token. 
The simulator enumerates all $K^T$ possible combinations of options and feeds them into the NL Modifier and the parser to get SQL queries. If one of these SQL queries is the same as the ground truth SQL, the simulator obtains the ideal selection. Otherwise, the PIIA fails to correct the NL question. Since it is too time-consuming to enumerate all $K^T$ combinations, we only rank and simulate the top 100. Concretely, we firstly filter out options whose tokens do not appear in the ground truth SQL. Then we provide a score to each left option, as Question Generator does in Section \ref{sec:questiongeneration}. Finally, the overall score for a combination is computed by summing up all the option scores in the combination. 

\begin{table}[t]
\centering
\scalebox{0.94}{
\label{tab:simulation_result_wikisql}
\begin{tabular}{l|lccc}
\toprule
\multicolumn{2}{c}{{\bf Models}} & \!\!\!\!\textbf{SQLAcc}& \!\textbf{ExeAcc}& \!\!\textbf{Avg.\#T} \!\!\!\\ 
\midrule
\multirow{4}{*}{\rotatebox{90}{WikiSQL}} 
&SQLova & 80.7 & 86.2 & \small{N/A} \\
&{~~~~~~+PIIA} & \textbf{84.9} & \textbf{88.9} & \textbf{1.3} \\
&SQLNet & 61.7 & 68.0 & \small{N/A} \\
&{~~~~~~+PIIA} & \textbf{68.4} & \textbf{73.2} & \textbf{1.7}  \\
\midrule
\multirow{6}{*}{\rotatebox{90}{Spider}} 
&IRNet  &  53.2 & \small{N/A} & \small{N/A} \\
&{~~~~~~+PIIA} &  \textbf{59.3} & \small{N/A} & \textbf{2.9} \\
&IRNet+BERT & 61.9 & \small{N/A} & \small{N/A} \\
&{~~~~~~+PIIA} & \textbf{63.4} & \small{N/A} & \textbf{2.4} \\
&SyntaxSQLNet\! &  27.2 & \small{N/A} & \small{N/A} \\
&{~~~~~~+PIIA} &  \textbf{34.2} & \small{N/A} & \textbf{3.4} \\
\bottomrule
\end{tabular}
}
\caption{Simulation results of PIIA on the WikiSQL test set and the Spider development set.}
\label{tab:simulation_result}
\end{table}

\paragraph{Model Comparison}

We evaluate PIIA with the simulator on both WikiSQL and Spider datasets. The results are shown in Table \ref{tab:simulation_result}. For the WikiSQL dataset, we report the accuracy of SQL exact matching (SQLAcc) and the accuracy of execution (ExeAcc). We can observe that PIIA boosts the performance for both base parsers with fewer than two average interaction turns (Avg.\#T). The absolute SQLAcc improvements for SQLova and SQLNet are 4.2\% and 6.7\% respectively, while the absolute ExeAcc improvements are 2.7\% and 5.2\%. The results on SQLNet are competitive with those of DailSQL \cite{gur2018dialsql}, a parser-independent method designed for simple SQL. However, on average, PIIA interacts with users by 1.7 turns, while DialSQL needs about 4.8. This indicates that PIIA is quite efficient and able to enhance text-to-SQL with only a few interaction turns.

Similar performance boosts can be observed on the Spider dataset, which has more complex multi-table SQL queries. As execution results are not available, we only report the results of SQLAcc. The improvements on IRNet, IRNet+BERT and SyntaxSQLNet again demonstrate the effectiveness of PIIA. PIIA can also enhance the parser integrated with BERT, which further proves the necessity of PIIA. 
After interacting with users, PIIA provides the revised NL questions in a form that is easier for parsers to understand. 
The average interaction turns are about three, a number users find acceptable. Fewer interaction turns are required by better parsers.
Additionally, a smaller improvement is obtained with a better parser, as better parsers know NL questions better. PIIA is effective because it utilizes the feedback provided by users, who know the questions best, thus leading to a further narrowing of the gap between parsers and users.

\subsection{Human Evaluation}

\begin{table}[t]
\centering
\scalebox{0.88}{
\begin{tabular}{lcccc}
\toprule
\!\!\!\textbf{Models} & \!\!\!\textbf{w/o PIIA} & \!\!\!\textbf{PIIA(H)} & \!\!\!\textbf{PIIA(S)} & \!\!\!\textbf{Avg.\#T}\\
\midrule
\!\!\!IRNet & 49.0 & 52.7 & 54.7 & 2.8 \\
\!\!\!IRNet+BERT & 60.7 & 62.2 & 62.7 & 2.4 \\
\bottomrule
\end{tabular}
}
\caption{SQL Accuracy of human evaluation (H) and simulation (S) on 300 samples.}
\label{tab:human}
\end{table}

We carry out the evaluation of PIIA with real users. The human evaluation is performed on the more complex Spider dataset and with two state-of-the-art parsers, i.e., IRNet and IRNet+BERT. We randomly sample 300 NL questions from the Spider development set and invite 30 volunteers majoring in liberal arts to interact with the PIIA agent. Each NL question is evaluated by three volunteers, all of whom are non-expert without any background knowledge of SQL queries. We provide them with the NL questions and the corresponding databases, and they interact with PIIA by answering the multi-choice questions.

The SQLAcc results of the human evaluation are shown in Table \ref{tab:human}. 
By interacting with real users, PIIA boosts the overall SQLAcc of both IRNet and IRNet+BERT by an absolute improvement of 3.7\% and 1.5\%, respectively. This indicates that PIIA provides a friendly way to interact with non-expert users, who are therefore able to understand the multi-choice questions and give proper answers. The average numbers of interaction turns are 2.8 and 2.4, respectively, which the users find acceptable. We also analyze the gap between human evaluation and simulation and find that some of the NL questions are ambiguous, making it hard for real users to distinguish similar options.

\subsection{Closer Analysis}

We use the IRNet parser on the Spider dataset to provide a closer analysis of PIIA. Similar observations can be obtained with other parsers.

\begin{figure}
\centering
\subfigure[]{\includegraphics[width=0.445\linewidth]{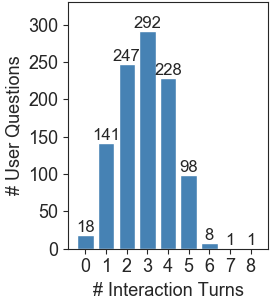}}
\hspace{2mm}
\subfigure[]{\includegraphics[width=0.445\linewidth]{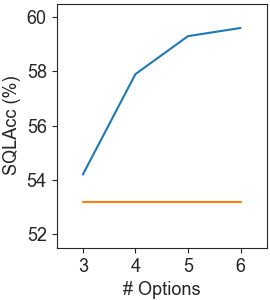}}
\caption{(a) Turn distribution and (b) SQLAcc w.r.t number of options by IRNet+PIIA on the Spider development set. Orange line in (b) indicates the result without PIIA.}
\label{fig:turns}
\end{figure}

\paragraph{Number of Turns}
In interactive systems, the number of interaction turns mostly determines whether users have a positive experience as too many turns may tire them out. With that in mind, we figure out the distribution of the number of turns (i.e., the number of multi-choice questions) over the Spider development set. The distribution is shown in Figure \ref{fig:turns}(a) with the average number being 2.9. 
As observed, the interaction process finishes in four turns in nearly 90\% of the cases. Only 10 out of 1,034 cases require an interaction process with more than five turns, which indicates PIIA is able to process such a complex dataset with high efficiency. 
We also analyze the cases correctly modified by the simulation and find that about 40\% of the multi-choice questions get the \emph{None} answer, which is an acceptable percentage. Additionally, the performance of our Error Locator is adequate. Though there are an average of 12.4 tokens in each NL question on the Spider development set, the Error Locator is able to find about three uncertain tokens out of them effectively.

\paragraph{Number of Options}
Providing a greater number of options in multi-choice questions increases the chances of including the correct one, thus enhancing the performance of the system. With more options, however, users have to make more effort. Therefore, we conduct experiments to analyze the influence of the number of options on the SQLAcc under simulation, as shown in Figure \ref{fig:turns}(b). The curve tends to be smooth after five options, meaning that this number is reasonable and able to balance the number of options and the need for correct ones. It's worth mentioning that each question contains two necessary options, i.e., \textit{None} and \textit{Value}, so that the least number of options is three.

\begin{figure}[t]
    \centering
    \includegraphics[width=.85\linewidth]{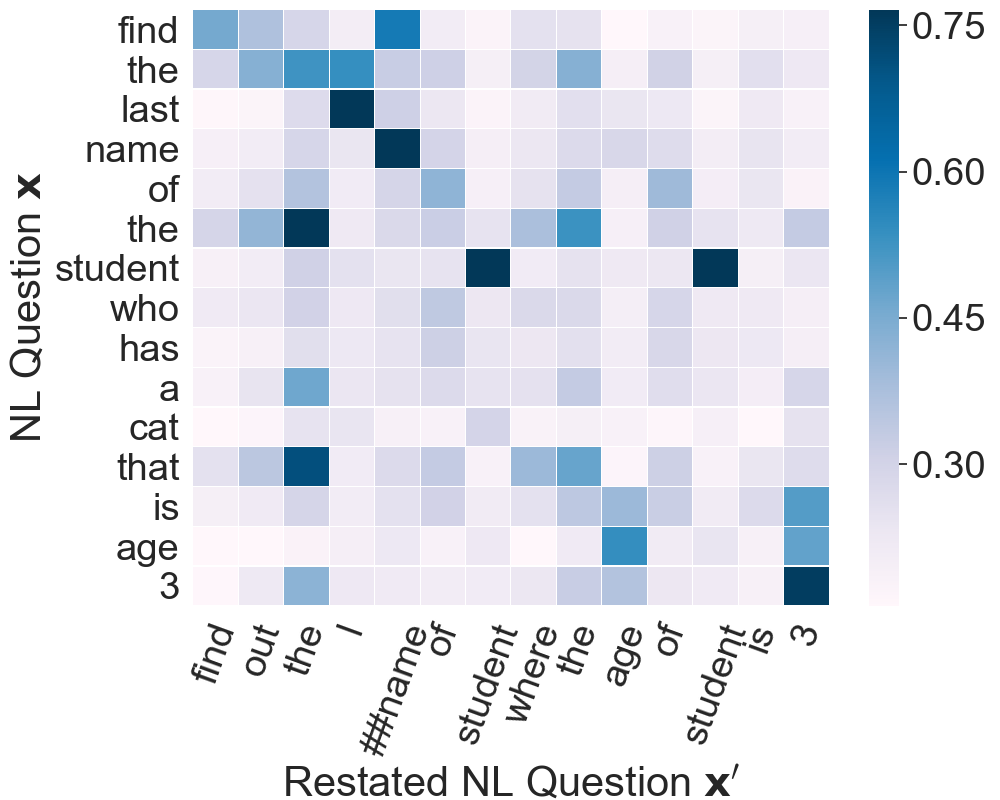}
    \caption{The similarity matrix of a real case.}
    \label{fig:similarity_matrix}
\end{figure}

\begin{table*}[t]
    \centering
        \begin{tabular}{c}
            %\hspace{-2.5mm}
            \includegraphics[width=0.97\textwidth]{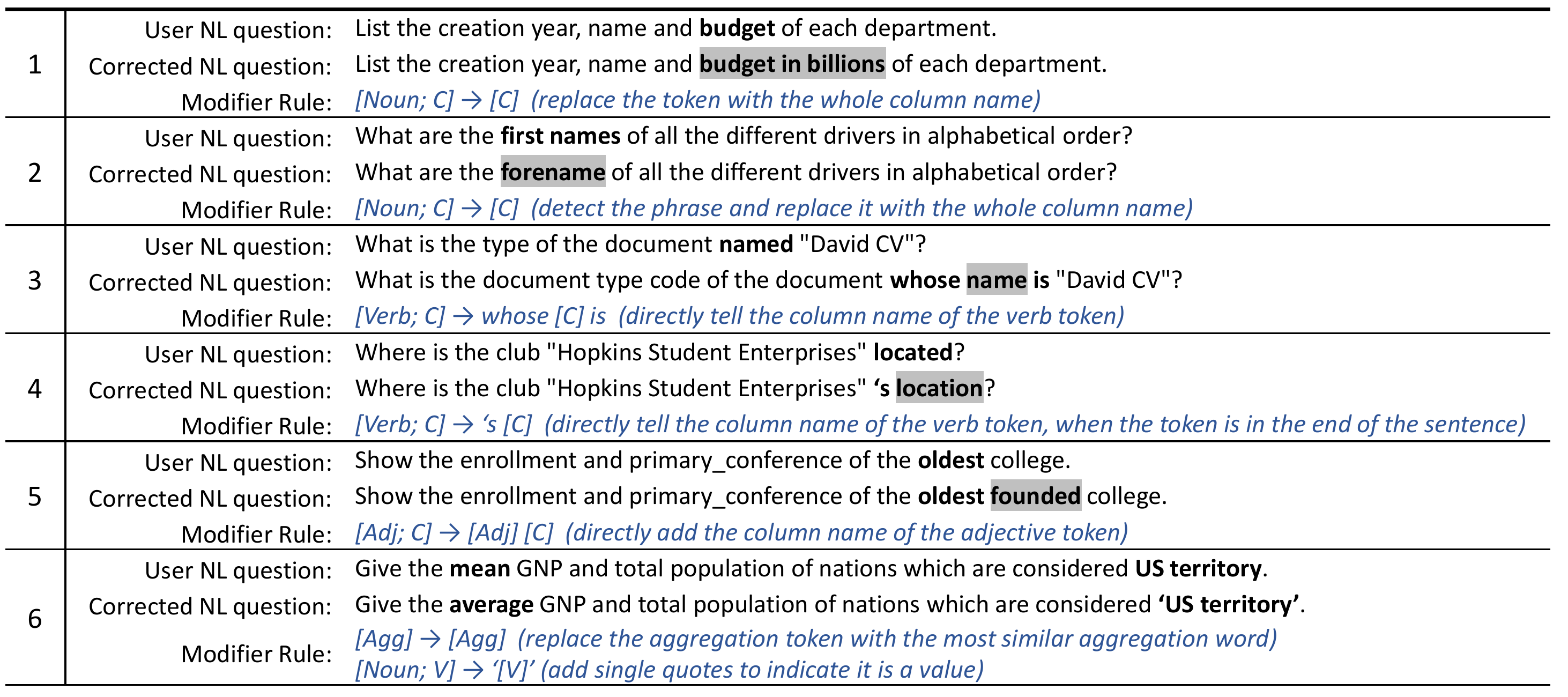}
        \end{tabular}
	\caption{Cases by IRNet+PIIA on Spider. Texts highlighted in gray indicate column names in the databases.}
	\label{fig:more_case}
\end{table*}

\paragraph{Similarity Matrix}

The Error Locator module is responsible for calculating a similarity matrix $A$ between a user's NL question $\mathbf{x}$ and a restated NL question $\mathbf{x}'$. Following the example in Figure \ref{fig:errorlocator}, we show the similarity matrix in Figure \ref{fig:similarity_matrix}. ``last name'' is evidently more similar to the column name ``lname'' than to the others, which meets our expectations. Two uncertain tokens are extracted, namely ``cat'' and ``age''. They are not stop words, and their scores don't reach the threshold. Since the token ``cat'' is not expressed in the restated NL question, it has a low similarity score. Although ``age'' appears in the restated NL question, it exists in two column names in the database, i.e., ``age'' from the table ``student'' and ``pet\_age'' from the table ``pet''. Thus the score of ``age'' is reduced and falls below the threshold.

\paragraph{Modifier Rules}
We carefully design the modifier rules for the NL Modifier based on the uncertain tokens and the selected options. Concretely, we take into consideration the types of options (column, table, value, or aggregation), the POS taggings of the uncertain tokens, and the contexts of the uncertain tokens. As shown in Table \ref{tab:simulation_result}, PIIA with the NL Modifier improves the efficacy of IRNet SQLAcc from 53.2\% to 59.3\%. Instead of using the NL Modifier, we try a straightforward way to modify the NL questions that involves directly replacing the uncertain tokens with selected options. With this approach, the simulated SQLAcc stands at 54.8\%, a result that is much worse than what PIIA can achieve with the NL Modifier, thus proving that our module is indispensable.

\paragraph{Case Study} 
Table \ref{fig:more_case} shows more real cases of users' NL questions and corrected NL questions along with the corresponding modifier rules. The words in bold are uncertain tokens and their corrections. Though IRNet wrongly parses these six cases, PIIA manages to solve them correctly. The first five cases are modified by rules for nouns, verbs, and adjectives that are related to the column names in the databases. Different rules are applied to add the column names into NL questions, making it more explicit for the parser to understand them. Case 6 shows an example of how to modify the aggregation operator and the value-related tokens. PIIA revises the inexplicit NL questions by interacting with users. Equipped with the PIIA agent, the performance of the base parser is improved.

\section{Related Works}

The works most related to ours are those investigating interactive semantic parsing. For instance, DailSQL, proposed by \citet{gur2018dialsql}, aims to detect error spans and their categories based on an encoder-decoder architecture. But it is designed for relatively simple scenarios. In this research area, another impressive work involves a model-based interaction system, which detects uncertain tokens and asks questions relying on inner parser states \cite{yao2019model}. Unlike these studies, however, we design a parser-independent interactive approach that can also perform cross-domain complex SQL queries. In the field of applied systems, \citet{gao2015datatone} focused on user interface designing and proposed an interactive semantic parsing system called Datatone. In contrast to them, our main contribution lies in the realm of technology. 
Another topic our method related to is query reformulation. The idea of query reformulation is explored by \citet{ray2018learning} and \citet{rastogi2019scaling}, while they apply this idea in other domains with different scenarios.
Our work is also related to semantic parsing, the process of converting natural language utterances into logical forms.
Sequence-to-sequence methods are widely applied to solve this task \cite{berant2013semantic,dong2016language,finegan2018improving,su2018natural}. To reduce search space for decoding, several works employed intermediate representations to generate abstract representations \cite{cheng2017learning,goldman2017weakly,dong2018coarse,qian2020how}. Although these methods have achieved an impressive performance in experimental studies, there is still a long way to go before they can be successfully applied in real systems.

Works dealing with the task of weakly supervised word alignment are also related to our research because our Error Locator module performs the same task. Some examples include the work of  
\citet{liu2015contrastive}, who proposed a latent-variable log-linear model for 
word alignment, the research of \citet{legrand2016neural}, who used pairwise training with negative sampling to train the alignment model, and a study that introduced a gradient-based alignment method for machine translation \cite{he2019towards}.

\section{Conclusion and Future Work}

We propose a parser-independent interactive approach, PIIA, to enhance the text-to-SQL process in NLIDB systems. PIIA interacts with users via multi-choice questions and can be built on arbitrary parsers. Experimental results show this approach leads to significant performance boosts on two cross-domain datasets with five different base parsers. In the future, we are interested in distilling and reusing the common knowledge from users' selections.

\section*{Acknowledgments}
We thank all the anonymous reviewers for their valuable comments. This work was supported in part by NSFC under Grant No. 61532001, National Key Research and Development Program of China under Grant No. 2018AAA0101902, and MOE-ChinaMobile Program under Grant No. MCM20170503.

\bibliography{emnlp2020}
\bibliographystyle{acl_natbib}

\end{document}